\documentclass{article}

    \PassOptionsToPackage{numbers, compress}{natbib}

\usepackage[preprint]{neurips_2026}

\usepackage[utf8]{inputenc} 
\usepackage[T1]{fontenc}    
\usepackage{hyperref}       
\usepackage{url}            
\usepackage{booktabs}       
\usepackage{amsfonts}       
\usepackage{nicefrac}       
\usepackage{microtype}      
\usepackage{xcolor}         
\usepackage{colortbl}       

\usepackage{amsmath}
\usepackage{graphicx}
\usepackage{subcaption}
\usepackage{multirow}
\usepackage{enumitem}
\usepackage{float}

\title{Predicting Decisions of AI Agents from Limited Interaction through Text-Tabular Modeling}

\author{%
  Eilam Shapira \quad Moshe Tennenholtz \quad Roi Reichart \\
  Faculty of Data and Decision Sciences\\
  Technion -- Israel Institute of Technology\\
  Haifa, Israel \\
  \texttt{\{eilam.shapira, moshe.tennenholtz, roireichart\}@gmail.com}
}

\begin{document}

\maketitle

\begin{abstract}
AI agents increasingly negotiate and transact in natural language with unfamiliar counterparts: a buyer bot facing an unknown seller, or a procurement assistant negotiating with a supplier. In such interactions, the counterpart's underlying LLM, prompts, control logic, and rule-based fallbacks are hidden, while each decision can have monetary consequences. We ask whether an agent can predict an unfamiliar counterpart's next decision from only a few prior interactions. To avoid real-world logging confounds, we study this problem in controlled bargaining and negotiation games, formulating it as target-adaptive text-tabular prediction: each decision point is a table row combining structured game state, offer history, and dialogue, while $K$ previous games of the same target agent, i.e., the counterpart being modeled, are provided in the prompt as labeled adaptation examples. Our model is built on a tabular foundation model that represents rows using game-state features and LLM-based text representations, and adds LLM-as-Observer as an additional representation: a small frozen LLM reads the public decision-time state and dialogue; its answer is discarded, and its hidden state becomes a decision-oriented feature, making the LLM an encoder rather than a direct few-shot predictor. Training on 13 frontier-LLM agents and testing on 91 held-out scaffolded agents, the full model outperforms direct LLM-as-Predictor prompting and game+text features baselines. Within this tabular model, Observer features contribute beyond the other feature schemes: at $K=16$, they improve response-prediction AUC by about 4 points across both tasks and reduce bargaining offer-prediction error by 14\%. These results show that formulating counterpart prediction as a target-adaptive text-tabular task enables effective adaptation, and that hidden LLM representations expose decision-relevant signals that direct prompting does not reliably surface.\footnote{Code and the $91$-agent dataset of $4{,}921$ bargaining and negotiation games will be released upon acceptance.}
\end{abstract}

\section{Introduction}
\label{sec:intro}

AI agents increasingly negotiate and transact in natural language with unfamiliar counterparts: a buyer bot facing an unknown seller, or a procurement assistant negotiating with a supplier. In such interactions, the counterpart’s underlying LLM, prompts, control logic, and rule-based fallbacks are hidden, while each decision can have monetary consequences. We ask whether an agent can predict an unfamiliar counterpart’s next decision from only a few prior interactions.

Real marketplace logs would be the most direct testbed, but they are rarely public and typically do not support systematic comparison across many agents under matched strategic conditions with known payoffs and ground-truth decisions. We therefore study the problem in controlled bargaining and negotiation games. These games preserve key elements of language-mediated commerce: multi-turn offers, accept/reject decisions, private valuations, monetary payoffs, and free-text dialogue. They also let us vary horizons, valuations, and information regimes while observing the decisions agents actually make.

\begin{figure*}[t]
  \centering
  \includegraphics[width=0.89\linewidth]{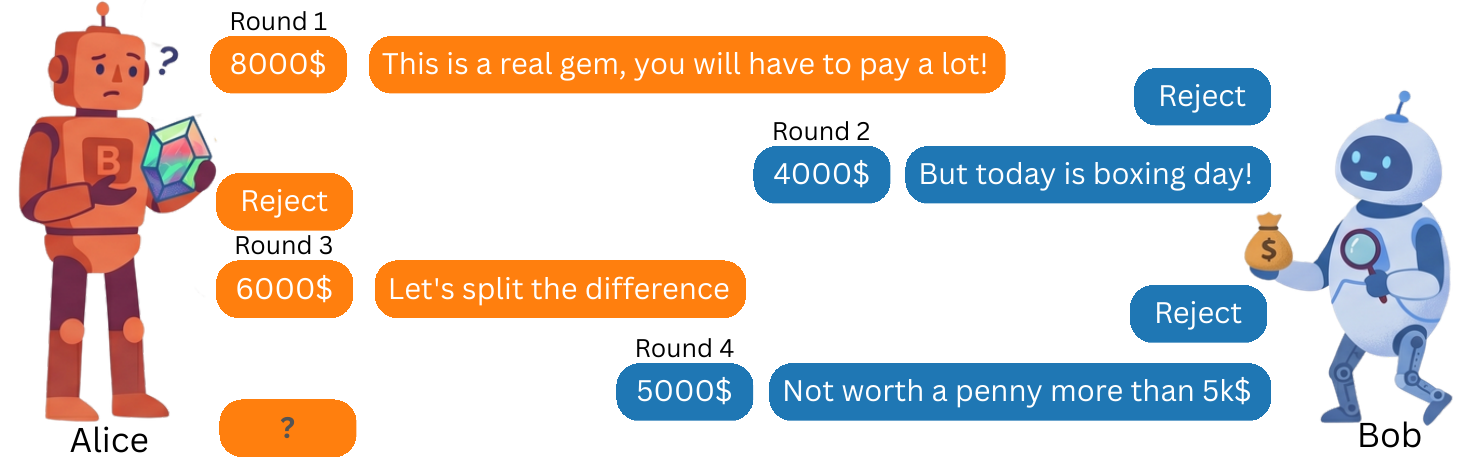}
  \caption{Alice (seller) and Bob (buyer) negotiate via free-text offers. Following Bob's \$5{,}000 round-4 offer, Alice's next move is the prediction target. \textbf{(a) Response prediction (classification):} will she accept? \textbf{(b) Proposal prediction (regression):} if she rejects, what will she propose?}
  \label{fig:setup}
\end{figure*}

We call the unfamiliar counterpart being modeled the \emph{target agent}. For each target, the predictor is given \(K\) complete prior games played by that same agent, which serve as labeled examples of the target’s behavior. At test time, the predictor receives a new decision point: the public game state, the offer history, and the dialogue so far. It must predict the target’s next move. We study two complementary tasks, illustrated in Figure~\ref{fig:setup}: \emph{response prediction}, a binary classification task asking whether the target accepts the current offer, and \emph{proposal prediction}, a regression task asking what offer the target will make next.

We formulate this as \emph{target-adaptive text-tabular prediction}. Each decision point is represented as a table row combining structured game variables, offer history, and dialogue-derived text features. A tabular foundation model conditions on labeled rows from a source population of previously observed agents together with the \(K\) labeled games of the current target agent. This allows the predictor to combine population-level regularities with target-specific evidence, adapting to a new counterpart without observing its prompt, code, or control logic.

Our model uses three complementary feature blocks. The first contains game-state features, such as public configuration variables, round number, current offer, and previous offers. The second contains generic text representations of the dialogue. The third is our new decision-oriented representation, \emph{LLM-as-Observer}: a small frozen LLM reads the public decision-time state and dialogue, its direct answer is discarded, and its hidden state is used as an additional feature for the tabular predictor. Thus, the LLM is used as an encoder rather than as the final few-shot predictor.

This design contrasts with a natural alternative, \emph{LLM-as-Predictor}: prompting a large frontier LLM with the current game and the target’s \(K\) prior games, and asking it to predict the next decision directly. Direct prompting can read the dialogue and reason over examples in context, but it must commit to an answer and cannot easily combine the target’s few games with a large labeled source population. In our formulation, the LLM contributes a reusable representation, while adaptation is performed by the tabular learner over source and target rows.

For the source population, we use the 13-agent round-robin tournament released as part of GLEE~\cite{shapira2024glee}, where frontier LLMs\footnote{frontier LLMs: state-of-the-art API Large Language Models} from six providers play under identical prompts, varying only in the underlying LLM. For the held-out target population, we introduce a 91-agent university-hackathon dataset: student-built agents that share one underlying LLM but differ in prompting, control logic, and rule-based fallbacks. This split tests whether predictors learned from one axis of agent variation transfer to newly encountered engineered agents whose heterogeneity comes from scaffolding.

The full target-adaptive text-tabular model, trained on 13 frontier-LLM agents and tested on 91 held-out scaffolded agents, outperforms direct LLM-as-Predictor prompting and game+text features baselines. Within the tabular model, Observer features add complementary signal beyond structured game features and generic dialogue representations. At \(K=16\), they improve response-prediction AUC by about four percentage points across both game families and reduce bargaining offer-prediction error by \(14\%\). The gain is not mainly in the Observer’s committed answer: hidden states provide substantially more value than its direct output, suggesting that frozen LLM representations expose decision-relevant information that direct prompting does not reliably surface.

\paragraph{Contributions.}
First, we formulate few-shot prediction of unfamiliar language-based agents as a target-adaptive text-tabular task, where \(K\) prior games of the target agent provide labeled adaptation examples. Second, we build a prediction model that combines game-state features, dialogue representations, and a new decision-oriented feature block, LLM-as-Observer. Third, we introduce a 91-agent hackathon dataset and a cross-population transfer evaluation from frontier-LLM agents to scaffolded agents, showing that the full model outperforms direct LLM-as-Predictor prompting and game+text features baselines, and that Observer hidden states add complementary decision-relevant signal.

\section{Related work}
\label{sec:related}

\paragraph{Multi-agent applications and the role of language.}
The applications motivating this paper sit in language-mediated commerce: consumer-to-consumer marketplaces~\citep{he2018decoupling, yang2021improving}, residential real-estate transactions~\citep{heddaya2023language}, tourism and travel-package negotiations~\citep{priya2025genteel}, multi-stakeholder contract deliberations~\citep{abdelnabi2023cooperation}, and the broader emerging ``agentic economy'' of LLM-based shopping and procurement assistants~\citep{rothschild2025agentic}, with early controlled deployments of LLM-vs-LLM marketplaces already reported~\citep{anthropic2026projectdeal}. They differ from non-language multi-agent AI such as multi-agent autonomous driving~\citep{cui2021scalable}, multi-robot coordination~\citep{escudie2024multisoc}, algorithmic trading~\citep{wang2025marketmaking}, and distributed power-grid control~\citep{chen2022powernet}, where agents observe each other through sensors, actions, and shared infrastructure, rather than through a dialogue. A second line of multi-agent learning research trains agents to coordinate through continuous vectors optimised end-to-end with their policies~\citep{sukhbaatar2016commnet} or through emergent discrete codes invented for the task~\citep{lazaridou2018emergence}: in those settings the communication channel is task-tuned, opaque to outside observers, and trained jointly with the policy. The setting we study sits on the opposite end of this axis: target agents emit fluent natural-language messages produced by pretrained LLMs~\citep{guo2024llmagents}, the channel itself is human-readable and not co-trained with the predictor, and any external observer must read the same public stream of strategic state and free-form dialogue that a human auditor would.

\paragraph{LLMs as strategic agents.}
A growing literature studies LLMs and other AI systems as strategic agents in language-mediated settings: bargaining and negotiation~\citep{shapira2024glee, xia2024measuring, kwon2024llms, bianchi2024negotiate}, persuasion and social influence~\citep{bao2026electwit, campedelli2024break, shapira2024can, shapira2025human, taubenfeld2024systematic}, auctions and market-like environments~\citep{chen2023aucarena, fish2024collusion, zheng2026marketbench}, social dilemmas and cooperation~\citep{lishirado2025spontaneous, backmann2025ethics, madmoun2025communication}, and broader social-agent benchmarks~\citep{zhou2024sotopia, xie2026m3bench, karten2025economist, wu2026malles}. Whereas this prior work characterises how LLMs behave as a population of strategic agents, we ask a per-agent predictive question: given $K$ observed games of a specific unseen agent, what will it decide next? Methods of population characterisation do not directly transfer to this task: they aggregate across agents, while we need to make a prediction at the individual-agent level.

\paragraph{Predicting agent behavior from limited histories.}
Predicting another actor's behaviour from limited interaction histories is a long-standing problem in multi-agent AI. Classical \emph{opponent-modelling} maintains beliefs over a library of hypothesised agent types and updates them from observed actions~\citep{albrecht2018autonomous, nashed2022survey, gmytrasiewicz2005framework, albrecht2013game, albrecht2016belief}; \emph{automated negotiation} learns preferences from partial dialogue~\citep{baarslag2016learning, coehoorn2004preferences, chawla2022opponent}; \emph{ad-hoc teamwork} predicts the behaviour of unfamiliar teammates~\citep{stone2010adhoc, mirsky2022survey, ribeiro2023teamster, wang2024naht}; and Theory-of-Mind networks~\citep{rabinowitz2018machine, nguyen2022traittom, nguyen2023memory, li2023tomllm, mu2026adaptive, xiao2025dyntom, zhang2025autotom} and predictors for human decisions in negotiation and persuasion~\citep{cadilhac2013grounding, shapira2024can, shapira2025human, shapira2026alignment, labruna2026winning} learn end-to-end from behavioural traces. These methods show that short histories can support prediction, but assume an agent type drawn from a known prior or a population matched to training, not an open-ended LLM-based agent whose implementation style is previously unseen. A modern alternative is to prompt a large API-based LLM in-context as a few-shot predictor~\citep{brown2020language, dong2024survey}. Throughout this paper we use ``LLM-as-Predictor'' to mean exactly this: a large API-based LLM prompted at inference time as a predictor. Our small-Observer pipeline is both cheaper at inference and more accurate (Section~\ref{sec:results}).

\paragraph{Multi-modal text--tabular learning.}
Each decision point in our setting combines structured game fields, such as offers,
round number, and configuration parameters, with free-form dialogue. We therefore
treat the task as text--tabular prediction. Tabular foundation models support
in-context prediction from labeled examples without gradient-based retraining
\cite{hollmann2023tabpfn,hollmann2025tabpfn,qu2025tabicl}, matching our
few-shot target-agent setting. Prior work studies text--tabular learning through
multi-modal AutoML, dedicated benchmarks, cross-table transfer, and foundation
models for tables with text fields
\cite{shi2021benchmarking,mraz2025benchmarking,arazi2026multabenchbenchmarkingmultimodaltabular,kim2024carte,
koloski2025llm,arazi2025tabstar}. Our setting differs in requiring rapid adaptation
to a newly observed strategic agent from only $K$ games, using source-population
rows and target-specific examples without gradient-based retraining.

\paragraph{Frozen LM representations as transferable features.}
Frozen LMs expose information through intermediate hidden states that is not
always captured by their final outputs. Probing work shows that syntactic,
semantic, and task-relevant variables can be decoded from these states
\cite{belinkov2022probing,conneau2018what,hewitt2019structural,tenney2019bert}. Related work
further shows that intermediate or layer-combined representations often transfer
better than final-layer outputs on downstream tasks
\cite{peters2018deep,hosseini2023bert,skean2025layer}. Recent studies also find that
hidden states can encode knowledge or signals that are not reflected in the
model's generated answer \cite{gekhman2025insideout,orgad2025llms}. We use this line
of work as motivation for a feature block in a text-tabular predictor: the
Observer reads the public game state and dialogue, but the downstream model
predicts the target agent's decision from its hidden state together with game and
dialogue features. This differs from standard probing in the target being
predicted: the representation is extracted from one model observing the
interaction, while the label is the next decision of another, black-box strategic
agent.

\section{Data}
\label{sec:data}

We instantiate our prediction task in GLEE~\citep{shapira2024glee}, a benchmark and simulation framework for two-player, sequential, language-based economic games. In GLEE, agents repeatedly make strategic decisions--such as proposing an offer or accepting/rejecting one--while observing the public interaction history and, in the language condition, exchanging free-text messages. The benchmark fixes the game rules while systematically varying payoff parameters, horizons, information regimes, and communication channels. This makes GLEE a natural source for our task: it preserves key ingredients of language-mediated commerce--private values, monetary incentives, multi-turn offers, and strategic dialogue--while providing controlled conditions and ground-truth agent decisions.

We focus on GLEE's two mixed-motive families most aligned with our prediction setting: bargaining and negotiation. In both, two agents alternate offers accompanied by free-text messages, and each decision point can be represented as a text-tabular row containing the public configuration, offer history, dialogue so far, and the target agent's next move. This yields our two prediction tasks: response prediction, asking whether the target accepts the current offer, and proposal prediction, asking what offer the target makes next.

\begin{table}[!b]
\centering
\footnotesize
\caption{Agent populations used for cross-population transfer, split by game family.}
\label{tab:populations}
\setlength{\tabcolsep}{3pt}
\begin{tabular}{llcccccc}
\toprule
Population & Family & \#~Agents & \#~Games & \#~Decisions & Avg.~\#~rounds & Variation axis & Role \\
\midrule
\multirow{2}{*}{GLEE tournament~\citep{shapira2024glee}}
 & Barg. & $13$ & $30{,}888$ & $97{,}031$ & $3.8$ & \multirow{2}{*}{LLM} & \multirow{2}{*}{source} \\
 & Nego. & $13$ & $30{,}940$ & $99{,}889$ & $4.0$ & & \\
\midrule
\multirow{2}{*}{University hackathon (ours)}
 & Barg. & $90$ & $2{,}936$ & $8{,}762$ & $6.0$ & \multirow{2}{*}{scaffolding} & \multirow{2}{*}{target} \\
 & Nego. & $60$ & $1{,}985$ & $2{,}579$ & $2.6$ & & \\
\bottomrule
\end{tabular}
\end{table}

\paragraph{Bargaining.}
Two agents divide a fixed sum $M$ over multiple rounds in an alternating-offers game~\citep{rubinstein1982perfect}. At each round, the proposer suggests a split $(p,1-p)$ and sends a message; the responder accepts, ending the game, or rejects, allowing the interaction to continue with reversed roles. Delay is costly through per-round discount factors $\delta_1,\delta_2 \in (0,1]$. Configurations vary in the horizon, the discount factors, and whether each agent observes the other's discount factor. Thus, agents must interpret both offers and language when deciding whether to concede, reject, or counter-offer.

\paragraph{Negotiation.}
A seller with private reserve value $V_S$ and a buyer with private valuation $V_B$ negotiate over the price of a single indivisible good. They alternate price offers, each accompanied by a free-text message. The responder can accept, ending the game; reject and continue when the horizon allows it; or exercise an outside option that guarantees zero surplus. For response prediction, we group outside-option decisions with rejection, since both are decisions not to accept the current offer. Configurations vary in the horizon, valuations, and whether each side observes the other's valuation. Because valuations are private, agents must infer value from offers, signal credibly through language, and decide when agreement remains worthwhile.

We use two complementary agent populations (Table~\ref{tab:populations}): the GLEE frontier-LLM tournament as the training source, where agents vary in the underlying LLM, and a new university-hackathon dataset as the held-out target population, where agents vary in scaffolding around a shared underlying LLM. This split tests whether predictors trained on one axis of agent variation transfer to newly encountered agents whose heterogeneity comes from a different source.

\paragraph{Frontier-LLM tournament (training source).}
The source population is the GLEE round-robin tournament: 13 frontier LLMs from six providers (full model list in Appendix~\ref{app:glee_models}) play bargaining and negotiation games under identical system prompts, so agents vary only in the underlying model. The tournament covers $960$ configurations over horizons, discount factors, valuations, information regimes, and communication regimes, yielding $\approx$64K games and 197K accept/reject decisions.

\paragraph{University hackathon (held-out target).}
The target population is a new dataset from a competitive university hackathon held in December 2025, where 34 teams competed for a \$2,000 prize. In contrast to the GLEE tournament, agents were restricted to the Gemini 2.5 Flash/Flash-Lite API surface but differed in scaffolding: engineered control logic, prompting pipelines, rule-based fallbacks, or combinations of these. We include logs from all competition stages, treating each submitted team-stage version as a distinct agent, yielding 91 agents, 4,921 games, and 11,341 decisions.

This source--target design tests whether predictors trained on agents that differ mainly in their underlying LLM transfer to agents that differ mainly in scaffolding.

\begin{figure*}[t]
  \centering
  \includegraphics[width=\linewidth]{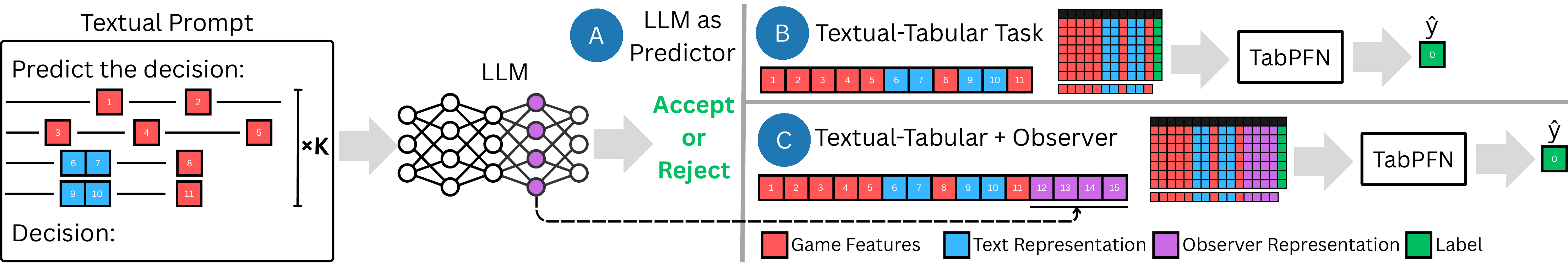}
  \caption{Three approaches for predicting decisions of a target agent. \textbf{(A) LLM-as-Predictor} receives the decision-time state, dialogue, and $K$ observed target games, and directly outputs the decision. \textbf{(B) Textual-tabular prediction} represents each decision point as a row of game features and dialogue. \textbf{(C) Our method} augments this row with Observer hidden-state representations from a frozen LLM.}
  \label{fig:methods}
\end{figure*}

\section{Method}
\label{sec:method}

Our goal is to predict the next decision of a previously unseen language-based agent from only a few observed games. The central design choice is to treat this as target-adaptive tabular prediction. Instead of asking an LLM to directly imitate the target agent, we represent each decision point through complementary feature modalities and let a tabular foundation model adapt to the target from its $K$ labeled games.

Figure~\ref{fig:methods} summarizes the model. At a decision point, the predictor observes only the public game state and the dialogue so far. We convert this information into three feature modalities: structured game-state features, a generic dialogue representation, and a decision-oriented hidden-state representation from a small frozen LLM, which we call the Observer. These features are combined by the same tabular predictor, which conditions on a large source population together with the target's $K$ observed games. We first define the prediction setting, then describe the three feature modalities, the tabular predictor, and the baselines.

\subsection{Prediction setting}

At each round, the target agent makes one of two types of decisions. In response prediction, the target receives an offer and must decide whether to accept it. This is a binary classification task. In proposal prediction, the target makes the next offer. This is a regression task over a normalized offer value. Together, these two tasks cover the main observable moves made by agents in bargaining and negotiation games.

For a new target agent, we are given $K$ previously observed games and must predict its decisions in held-out games. The target itself is never queried at inference time, and we never observe its prompt, code, or control logic. All predictors receive only the information that would be public at the decision point. In private-information configurations, values that are private to either player are masked and are not supplied to the game+text features, the LLM-as-Predictor prompt, or the Observer input.

\subsection{Feature modalities}

Each decision point is represented by three complementary modalities: structured game-state features, a generic dialogue representation, and the Observer hidden-state representation (Figure~\ref{fig:feature_modalities}). Together they form a single multimodal tabular row that the predictor of Section~\ref{sec:method} consumes.

\begin{itemize}[leftmargin=1.2em, nosep]
\item \textbf{Game-state features.} These features encode the structured strategic state of the game: the public configuration, the current offer, the round index, previous offers and decisions, and negotiation-specific information such as outside options when they are public. This modality gives the predictor direct access to the incentives and history that shape rational play.
\item \textbf{Dialogue representation.} Because agents communicate in natural language, the same offer can have different implications depending on the accompanying message. We therefore encode the dialogue so far with a sentence encoder and reduce the representation before passing it to the tabular predictor. This modality captures semantic information from the conversation, but it is not explicitly trained to represent the target's strategic decision.
\item \textbf{Observer representation.} The Observer is a small frozen LLM that reads the public decision-time state and dialogue. It is prompted toward the same decision the target is about to make, but its direct answer is discarded. Instead, we extract an internal hidden state and use it as a decision-oriented representation of the situation. The Observer is never fine-tuned, never sees the target's prompt or code, and does not receive the target's $K$ past games in its prompt. Adaptation to the target happens only in the downstream tabular predictor.
This separation is a key methodological observation. A frontier LLM-as-Predictor must both understand the situation and directly commit to a prediction. LLM-as-Observer uses the LLM only for the first role: constructing a representation. The final decision is made by a tabular model that can combine this representation with source-population data and the target's few labeled games.
\end{itemize}

Because the conditioning set mixes rows from many source agents with rows from the current target agent, identical game states may correspond to different behavioral policies. The agent-identity indicator tells the tabular predictor which rows come from the same decision-maker, allowing it to distinguish population-level regularities from target-specific deviations. Without this marker, source and target rows are treated as exchangeable observations even when they are generated by different agents.

\subsection{Tabular predictor}

The prediction module performs target-adaptive tabular inference over the multimodal row representation. For each target, the predictor conditions on labeled examples from the source population together with the target's $K$ observed games. The same predictor is used for both tasks: classification mode for response prediction and regression mode for proposal prediction.

\subsection{Baselines and controls}

We evaluate the full target-adaptive text-tabular model against direct prompting and reduced tabular baselines, and then isolate the marginal contribution of the Observer feature block.

\begin{itemize}[leftmargin=1.2em, nosep]
\item \textbf{Game+text features baseline.} This baseline uses the same tabular predictor and the same target-adaptation protocol, but removes the Observer representation. It receives only the structured game-state features, the dialogue representation, and the agent-identity indicator. This tests whether the Observer adds information beyond a strong tabular model.
\item \textbf{LLM-as-Predictor.} This baseline prompts a frontier LLM with the current game, the dialogue, and the target's $K$ observed games, and asks it to predict the target's next decision directly. This tests the natural alternative of using a large LLM as the predictor itself. The approach can read the dialogue and reason over examples in context, but it must commit to a prediction from the prompt alone and does not produce a reusable representation that can be combined with labeled source-population rows.
\end{itemize}

\section{Experimental setup}
\label{sec:setup}

\paragraph{Evaluation protocol.}
Our main evaluation is cross-population transfer. We train on the frontier-LLM tournament population and test on held-out hackathon agents, one target at a time. The source population varies the underlying LLM while holding scaffolding fixed; the target population holds the underlying LLM fixed while varying prompts, control logic, and rule-based fallbacks. This protocol tests whether features learned from one axis of agent variation transfer to the other.

For each target, we sample $K \in \{0, 2, 4, 8, 16\}$ observed games as adaptation examples and evaluate on the remaining games. Response prediction is evaluated with AUC, and proposal prediction is evaluated with $R^{2}$ over normalized offers. The tabular classifier is trained on up to $3{,}000$ individual decisions drawn from GLEE source agents (balanced across agents) together with the target's $K$-game decisions. We instantiate the tabular prediction module with TabPFN~v2.6. Observer metrics average over each model's upper-stack layer band (relative depth $0.6$--$0.9$), removing the per-layer optimisation knob. Observer models are small frozen LLMs (Gemma-2-2B, Qwen3-1.7B, and Llama-3.2-1B; 1--2B parameters). The LLM-as-Predictor baseline uses Gemini~2.5~Flash--the model made available to hackathon participants--giving direct prompting a substantial capacity advantage over the Observer.

\section{Results}
\label{sec:results}

We evaluate cross-population transfer from the $13$-agent frontier-LLM tournament to the held-out hackathon agents. The results are organized around two questions. First, does a target-adaptive tabular predictor outperform direct LLM prompting? Second, does the Observer hidden state add information beyond game+text features? All tabular methods use the same TabPFN predictor, with \emph{the Observer} referring to the configuration using game features, text embeddings, and LLM hidden states. Specific models are named by their LLM (e.g., \emph{Gemma-as-Observer}), while \emph{Game+text features} is the baseline without hidden states. The only non-tabular method is \emph{LLM-as-Predictor}, which prompts a frontier API directly. Throughout, $K$ is the number of adaptation games, and performance is measured by AUC (response) and $R^{2}$ (proposals).

\begin{table}[t]
  \centering
  \caption{Cross-population transfer. \textbf{Response} (left): mean AUC, $5$ seeds; For Observer methods, we average results over layers at relative depth $0.6$--$0.9$. \textbf{Proposal} (right): median $R^{2}$ on scale-normalised offers, $5$ seeds. $\pm$SE. \textbf{Bold}: best per family per $K$; \colorbox{gray!20}{shading}\!: Observer beats strongest baseline.}
  \label{tab:results_response}
  \label{tab:results_proposal}
  \small
  \setlength{\tabcolsep}{3pt}
  \resizebox{\textwidth}{!}{%
  \begin{tabular}{@{}c l ccccc|ccccc@{}}
    \toprule
    & & \multicolumn{5}{c}{\emph{Response (AUC)}} & \multicolumn{5}{c}{\emph{Proposal ($R^{2}$)}} \\
    \cmidrule(lr){3-7} \cmidrule(l){8-12}
    & Method & $K{=}0$ & $K{=}2$ & $K{=}4$ & $K{=}8$ & $K{=}16$ & $K{=}0$ & $K{=}2$ & $K{=}4$ & $K{=}8$ & $K{=}16$ \\
    \midrule
    \multirow{5}{*}{\rotatebox[origin=c]{90}{Barg.}}
    & Game+text features           & .621\,\scriptsize{$\pm$.01} & .683\,\scriptsize{$\pm$.01} & .712\,\scriptsize{$\pm$.01} & .748\,\scriptsize{$\pm$.01} & .791\,\scriptsize{$\pm$.01} & .366\,\scriptsize{$\pm$.08} & .407\,\scriptsize{$\pm$.10} & .503\,\scriptsize{$\pm$.08} & .502\,\scriptsize{$\pm$.07} & .622\,\scriptsize{$\pm$.05} \\
    & LLM-as-Predictor                  & .577\,\scriptsize{$\pm$.01} & .686\,\scriptsize{$\pm$.01} & .726\,\scriptsize{$\pm$.01} & .750\,\scriptsize{$\pm$.01} & .770\,\scriptsize{$\pm$.01} & $-2.8$\,\scriptsize{$\pm$.80} & $-1.6$\,\scriptsize{$\pm$.24} & $-0.5$\,\scriptsize{$\pm$.19} & $-0.5$\,\scriptsize{$\pm$.19} & $-0.3$\,\scriptsize{$\pm$.27} \\
    & Gemma-as-Observer             & \cellcolor{gray!20}\textbf{.666}\,\scriptsize{$\pm$.01} & \cellcolor{gray!20}\textbf{.719}\,\scriptsize{$\pm$.01} & \cellcolor{gray!20}\textbf{.754}\,\scriptsize{$\pm$.01} & \cellcolor{gray!20}\textbf{.793}\,\scriptsize{$\pm$.01} & \cellcolor{gray!20}\textbf{.831}\,\scriptsize{$\pm$.01} & \cellcolor{gray!20}.470\,\scriptsize{$\pm$.07} & \cellcolor{gray!20}.513\,\scriptsize{$\pm$.07} & \cellcolor{gray!20}\textbf{.598}\,\scriptsize{$\pm$.06} & \cellcolor{gray!20}.635\,\scriptsize{$\pm$.05} & \cellcolor{gray!20}\textbf{.676}\,\scriptsize{$\pm$.03} \\
    & Qwen3-as-Observer             & \cellcolor{gray!20}.655\,\scriptsize{$\pm$.01} & \cellcolor{gray!20}.714\,\scriptsize{$\pm$.01} & \cellcolor{gray!20}.749\,\scriptsize{$\pm$.01} & \cellcolor{gray!20}.786\,\scriptsize{$\pm$.01} & \cellcolor{gray!20}.826\,\scriptsize{$\pm$.01} & \cellcolor{gray!20}\textbf{.506}\,\scriptsize{$\pm$.06} & \cellcolor{gray!20}\textbf{.530}\,\scriptsize{$\pm$.07} & \cellcolor{gray!20}.590\,\scriptsize{$\pm$.05} & \cellcolor{gray!20}\textbf{.641}\,\scriptsize{$\pm$.04} & \cellcolor{gray!20}.673\,\scriptsize{$\pm$.02} \\
    & Llama-as-Observer             & \cellcolor{gray!20}.642\,\scriptsize{$\pm$.01} & \cellcolor{gray!20}.695\,\scriptsize{$\pm$.01} & \cellcolor{gray!20}.730\,\scriptsize{$\pm$.01} & \cellcolor{gray!20}.770\,\scriptsize{$\pm$.01} & \cellcolor{gray!20}.812\,\scriptsize{$\pm$.01} & \cellcolor{gray!20}.486\,\scriptsize{$\pm$.07} & \cellcolor{gray!20}.519\,\scriptsize{$\pm$.07} & \cellcolor{gray!20}.596\,\scriptsize{$\pm$.06} & \cellcolor{gray!20}.638\,\scriptsize{$\pm$.04} & \cellcolor{gray!20}.674\,\scriptsize{$\pm$.03} \\
    \midrule
    \multirow{5}{*}{\rotatebox[origin=c]{90}{Nego.}}
    & Game+text features           & .719\,\scriptsize{$\pm$.01} & .768\,\scriptsize{$\pm$.01} & .765\,\scriptsize{$\pm$.01} & .767\,\scriptsize{$\pm$.01} & .803\,\scriptsize{$\pm$.01} & .552\,\scriptsize{$\pm$.11} & .579\,\scriptsize{$\pm$.12} & .567\,\scriptsize{$\pm$.11} & .723\,\scriptsize{$\pm$.12} & .857\,\scriptsize{$\pm$.08} \\
    & LLM-as-Predictor                  & .486\,\scriptsize{$\pm$.01} & .684\,\scriptsize{$\pm$.01} & .684\,\scriptsize{$\pm$.01} & .713\,\scriptsize{$\pm$.01} & .785\,\scriptsize{$\pm$.02} & $-8.0$\,\scriptsize{$\pm$4.8} & $-3.2$\,\scriptsize{$\pm$2.9} & $-0.4$\,\scriptsize{$\pm$.81} & .164\,\scriptsize{$\pm$.21} & .506\,\scriptsize{$\pm$.29} \\
    & Gemma-as-Observer             & \cellcolor{gray!20}.738\,\scriptsize{$\pm$.01} & \cellcolor{gray!20}.782\,\scriptsize{$\pm$.01} & \cellcolor{gray!20}.794\,\scriptsize{$\pm$.01} & \cellcolor{gray!20}.792\,\scriptsize{$\pm$.01} & \cellcolor{gray!20}.826\,\scriptsize{$\pm$.01} & .551\,\scriptsize{$\pm$.12} & .578\,\scriptsize{$\pm$.13} & \cellcolor{gray!20}\textbf{.583}\,\scriptsize{$\pm$.12} & \cellcolor{gray!20}\textbf{.731}\,\scriptsize{$\pm$.12} & .845\,\scriptsize{$\pm$.09} \\
    & Qwen3-as-Observer             & \cellcolor{gray!20}\textbf{.808}\,\scriptsize{$\pm$.01} & \cellcolor{gray!20}\textbf{.825}\,\scriptsize{$\pm$.01} & \cellcolor{gray!20}\textbf{.833}\,\scriptsize{$\pm$.01} & \cellcolor{gray!20}\textbf{.833}\,\scriptsize{$\pm$.01} & \cellcolor{gray!20}\textbf{.852}\,\scriptsize{$\pm$.01} & .540\,\scriptsize{$\pm$.12} & .573\,\scriptsize{$\pm$.13} & .558\,\scriptsize{$\pm$.12} & .711\,\scriptsize{$\pm$.11} & .852\,\scriptsize{$\pm$.09} \\
    & Llama-as-Observer             & \cellcolor{gray!20}.743\,\scriptsize{$\pm$.01} & \cellcolor{gray!20}.774\,\scriptsize{$\pm$.01} & \cellcolor{gray!20}.782\,\scriptsize{$\pm$.01} & \cellcolor{gray!20}.779\,\scriptsize{$\pm$.01} & \cellcolor{gray!20}.810\,\scriptsize{$\pm$.01} & \cellcolor{gray!20}\textbf{.555}\,\scriptsize{$\pm$.12} & \cellcolor{gray!20}\textbf{.582}\,\scriptsize{$\pm$.13} & .544\,\scriptsize{$\pm$.11} & \cellcolor{gray!20}.725\,\scriptsize{$\pm$.12} & \cellcolor{gray!20}\textbf{.862}\,\scriptsize{$\pm$.09} \\
    \bottomrule
  \end{tabular}%
  }
\end{table}

\paragraph{Response prediction: Observer hidden states are the strongest predictor.}
Table~\ref{tab:results_response} shows that LLM-as-Observer improves response prediction across both game families and all values of $K$. In bargaining, the best Observer yields a substantial gain of $+4.0$pp over the game+text features baseline and $+6.1$pp over the LLM-as-Predictor at $K{=}16$. In negotiation, the best Observer provides a $+4.9$pp improvement over game+text features and $+6.7$pp over LLM-as-Predictor. The same pattern appears already at $K{=}0$, where the Observer improves over the tabular baseline without any target-specific examples.

\paragraph{LLM-as-Predictor is weaker.}
The LLM-as-Predictor baseline is a demanding comparison: it uses a large frontier API model, receives the current game and the target's $K$ observed games directly in context, and is from the same model family used by most hackathon agents. Nevertheless, Table~\ref{tab:results_response} shows that it trails LLM-as-Observer at every $K$ in both families. The failure is not due to the Predictor lacking language understanding or being mismatched to the target population. Direct few-shot prompting is a weaker interface for this prediction problem than extracting a reusable representation and letting the tabular model adapt over labeled source and target rows.

\paragraph{Proposal prediction: the Observer helps when structured history is not enough.}
Table~\ref{tab:results_proposal} shows a more nuanced effect for proposal prediction. In bargaining, all three Observer variants improve over game+text features across $K$, yielding a median $R^{2}$ increase of approximately $+0.05$ over the game+text features baseline at $K{=}16$. Using the Gemma-2-2B Observer as a representative point, this reduces the typical one-offer prediction error on a nominally \$$10{,}000$ split from \$$552$ to \$$473$ at $K{=}16$, a $14\%$ reduction. Importantly, this gain is specific to the bargaining setting, where the strategic dynamic relies heavily on interpreting text alongside numerical offers. In negotiation, by contrast, Table~\ref{tab:results_proposal} shows that the game+text features baseline is already very strong at $K{=}16$, and the Observer variants do not provide a clear additional improvement. The pattern in Table~\ref{tab:results_proposal} therefore sharpens the claim: Observer hidden states help when the next offer is not already captured by the structured game history.

\paragraph{LLM-as-Predictor is especially weak for numerical offers.}
Table~\ref{tab:results_proposal} also shows that direct LLM prompting is poorly calibrated for proposal prediction. In bargaining, LLM-as-Predictor has a negative median $R^{2}$ even at $K{=}16$. In negotiation, it improves with more examples but still does not match the game+text features baseline. The large LLM can read the game and produce plausible numbers, but it is not a reliable regression model: autoregressive token decoding is poorly suited to calibrated numerical regression, and the in-context $K$-shot mechanism that helps the Predictor on binary classification has weak traction on continuous values. The tabular formulation is therefore not merely cheaper or more convenient; it is the right prediction interface for turning few observed games into calibrated numerical estimates.

\section{Robustness and ablation}
\label{sec:analysis}

\begin{table}[t]
  \centering
  \caption{Feature ablation at $K{=}16$ (Gemma-2-2B Observer). Left: leave-one-out from the full model; Right: reduced feature stacks. \textbf{G}~=~Game, \textbf{T}~=~Text, \textbf{O}~=~Observer, \textbf{I}~=~Identity. Results show mean AUC (response) and median $R^{2}$ (proposal), averaged over mid-to-late Observer layers.}
  \label{tab:ablation}
  \footnotesize
  \setlength{\tabcolsep}{4pt}
  \begin{tabular}{@{}ll ccccc cccc@{}}
    \toprule
    & & \textbf{Full} & $-$O & $-$T & $-$G & $-$I & G$+$I & T$+$I & O$+$I & I \\
    \midrule
    \multirow{2}{*}{Response (AUC)} & Barg. & .831 & .791 & .833 & .778 & .779 & .796 & .583 & .783 & .500 \\
                                    & Nego. & .826 & .803 & .815 & .615 & .817 & .779 & .621 & .604 & .500 \\
    \midrule
    \multirow{2}{*}{Proposal ($R^{2}$)} & Barg. & .689 & .616 & .695 & .618 & .607 & .621 & .013 & .625 & $-$.038 \\
                                        & Nego. & .841 & .844 & .843 & .450 & .722 & .843 & .063 & .438 & $-$.057 \\
    \bottomrule
  \end{tabular}
\end{table}

We isolate the load-bearing components of our framework to validate its two primary pillars: the text-tabular formulation and the integration of frozen Observer representations.

\paragraph{The feature hierarchy.}
\label{sec:analysis:ablation}
Table~\ref{tab:ablation} reveals a clear performance hierarchy across four feature blocks. Structured game features provide the essential backbone; removing them leads to the most substantial performance collapse, particularly in negotiation. The Observer hidden states supply the critical situational layer that generic dialogue embeddings fail to capture. Notably, once the Observer is integrated, generic sentence embeddings become largely redundant, providing no meaningful marginal gains.

\paragraph{Latent representation vs.\ direct prediction.}
A key architectural choice is using the LLM as an Observer rather than a Predictor. Analysis shows that feeding the hidden states into the tabular model consistently outperforms the Observer's direct accept/reject probabilities (logits) across all conditions (Appendix~\ref{app:providers}). This justifies the text-tabular approach: the tabular learner (TabPFN) decodes strategic signals from the LLM's latent space more effectively than the LLM's own output head.

\begin{figure}[t]
  \centering
  \includegraphics[width=\linewidth]{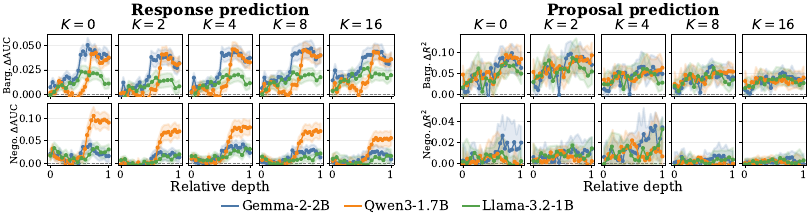}
  \caption{Observer gain over the game+text features baseline by relative depth. Observer gains are stable across mid-to-late layers (relative depth $0.6$--$0.9$) (\textbf{Left}: Response, $\Delta$AUC; \textbf{Right}: Proposal, $\Delta R^{2}$). Rows: bargaining (top), negotiation (bottom); columns: K-shot examples.}
  \label{fig:observer_layer}
\end{figure}

\paragraph{Stability across providers, tasks, and layers.}
\label{sec:results:robustness}
The Observer effect is robust across encoder LLMs: hidden states from Gemma-2-2B, Qwen3-1.7B, and Llama-3.2-1B all yield a consistent improvement over the baseline when fed into the tabular model. Figure~\ref{fig:observer_layer} explains this stability and shows why the hidden state is the relevant LLM signal: rather than peaking at a single tuned layer, the gains remain stable across mid-to-late layers (relative depth $0.6$--$0.9$) across both providers and the response and bargaining-proposal tasks. This confirms that the predictive signal is a stable, intrinsic property of mid-to-late Observer representations rather than an artifact of specific layer selection.

\section{Discussion and conclusion}
\label{sec:discussion}

The main takeaway of this paper is that predicting the decisions of an unfamiliar language-based agent is better framed as target-adaptive text-tabular learning than as direct few-shot LLM prediction. In this formulation, structured game-state and offer-history features provide the strategic backbone, dialogue features expose the language channel, and the target's K observed games provide adaptation evidence. The LLM-as-Observer adds a decision-oriented feature block that substantially improves the model.

This framing explains why the text-tabular model outperforms direct prompting. A frontier LLM-as-Predictor can read the current interaction and the target’s past games, but it must compress all evidence into a single generated answer and cannot naturally combine those examples with a large labeled source population. The tabular learner instead conditions jointly on source-population rows and target-specific rows, which better matches the statistical structure of the problem.

Within our model, the Observer adds a complementary signal. It consistently improves response prediction across both game families and outperforms LLM prompting despite using much smaller frozen LLMs. The gains do not come mainly from the Observer's final answer, but from its hidden representation, suggesting that frozen LLMs encode decision-relevant information that is not reliably exposed in their generated output. For proposal prediction, the picture is more nuanced: the Observer helps in bargaining, where language and strategic positioning matter beyond offer history, but adds little in negotiation, where the next offer is already well predicted from the structured game state, offer history, and target examples.

Our cross-population evaluation tests transfer to deployment-like agents. The source population consists of controlled frontier-LLM agents, which provide a broad labeled prior over strategic behavior. The held-out hackathon population consists of black-box scaffolded agents that differ in prompting, control logic, and rule-based fallbacks. Training on the former and testing on the latter evaluates whether a predictor learned from a reusable controlled source population can adapt to newly encountered engineered agents.

While our results are encouraging, several limitations remain. The games are controlled abstractions of language-mediated commerce, not real markets; the method assumes access to a relevant source population; and the Observer's contribution varies across tasks. Overall, our results suggest a general recipe for modeling the decision making of unfamiliar AI agents: separate representation from adaptation. Use language models to construct decision-relevant representations of strategic dialogue, but let the final prediction be made by a supervised model that can combine structured incentives, source-population evidence, and the target's few observed decisions.

\begin{ack}
Eilam Shapira is supported by a Google PhD Fellowship. Roi Reichart has been partially supported by a VATAT grant on data science. We thank Alan Arazi, Maya Zadok, and Shoham Grunblat for helpful comments on earlier versions of this work. We are grateful to Itamar Reichman, Gur Keinan, Idan Hahn, Gila Molcho, Omer Ben Porat, Avigdor Gal, and Rann Smorodinsky for their support in organizing the hackathon.
\end{ack}

\bibliographystyle{plainnat}
\bibliography{references}

\appendix

\section{Game configurations}
\label{app:configs}

This section lists the exact game configurations played by each population, in support of Section~\ref{sec:data}. The hackathon target population is restricted to configurations with free-text messages enabled (\texttt{messages\_allowed}=true); the GLEE source population includes both communication regimes.

\paragraph{GLEE source population.}
The frontier-LLM tournament sweeps $384$ distinct bargaining configurations and $576$ distinct negotiation configurations. Bargaining varies the money-to-divide $M \in \{100, 10\,000, 1{,}000{,}000\}$, the round horizon $\text{max\_rounds} \in \{12, \infty\}$, the information regime (complete vs.\ incomplete), the communication regime (free-text messages enabled vs.\ disabled), and the per-player discount factors $\delta_1, \delta_2 \in \{0.8, 0.9, 0.95, 1.0\}$. Negotiation varies the seller's reserve value $V_S$ and the buyer's valuation $V_B$ (each on a relative scale, with both values $\in \{0.8, 1.0, 1.2, 1.5\}$ scaled by a price order $\in \{100, 10\,000, 1{,}000{,}000\}$), the round horizon $\text{max\_rounds} \in \{1, 10, \infty\}$, the information regime, and the communication regime.

\paragraph{University-hackathon target population.}
A hackathon configuration uses the same GLEE parameterisation as the source population: in addition to the headline parameters released to participants in advance (horizon, information regime, parameter ranges), each game also fixes per-player payoff parameters that the engine uses to evaluate outcomes — the discount factors $\delta_1, \delta_2$ in bargaining and the valuations $V_S, V_B$ in negotiation. The specific values used at evaluation time were not disclosed to teams in advance, so they could not be hard-coded into agent design. At runtime, each agent always observes its own parameter (its own $\delta$ in bargaining, its own valuation in negotiation); the opponent's parameter is observed only under the complete-information regime, and is private otherwise — exactly as in GLEE. All such parameters are recorded in each game's \texttt{config.json} and vary across configurations within a stage. We therefore enumerate configurations on the full GLEE parameter set, which yields $10$ distinct bargaining and $8$ distinct negotiation configurations across the four stages. Table~\ref{tab:hackathon_configs} gives the full list.

\begin{table}[h]
  \centering
  \small
  \caption{Configurations played at each hackathon stage, enumerated on the full GLEE parameter set (so per-player discount factors $\delta_1,\delta_2$ that vary within a stage produce distinct configurations). \textbf{Bargaining columns:} $M$ = money to divide; max R = max rounds; info = complete~(C)~or~incomplete~(I) information; $\delta_1,\delta_2$ = per-player discount factors. \textbf{Negotiation columns:} $V_S, V_B$ = seller / buyer relative values; price order = scale; max R, info as above (negotiation configurations do not parameterise per-player discounting). All configurations have $\texttt{messages\_allowed}=\text{true}$.}
  \label{tab:hackathon_configs}
  \begin{tabular}{@{}l c c c c c c c c c c@{}}
    \toprule
    \multirow{2}{*}{Stage} & \multicolumn{5}{c}{Bargaining} & & \multicolumn{4}{c}{Negotiation} \\
    \cmidrule(lr){2-6} \cmidrule(l){8-11}
    & $M$ & max R & info & $\delta_1$ & $\delta_2$ & & $V_S, V_B$ & price order & max R & info \\
    \midrule
    \multirow{3}{*}{1}
      & \multirow{3}{*}{$100$} & \multirow{3}{*}{$12$} & \multirow{3}{*}{C} & $0.8$  & $0.95$ & & --- & --- & --- & --- \\
      & & & & $0.8$  & $1.0$  & & --- & --- & --- & --- \\
      & & & & $0.95$ & $0.95$ & & --- & --- & --- & --- \\
    \midrule
    \multirow{2}{*}{2}
      & \multirow{2}{*}{$10{,}000$} & \multirow{2}{*}{$12$} & \multirow{2}{*}{I} & \multirow{2}{*}{$0.8$} & \multirow{2}{*}{$0.8$}
        & & $1.0,\ 1.2$ & $10{,}000$ & $1$ & C \\
      & & & & & & & $0.8,\ 1.5$ & $10{,}000$ & $1$ & I \\
    \midrule
    3 & $1{,}000{,}000$ & $12$ & I & $0.9$ & $0.9$ & & $1.0,\ 1.5$ & $1{,}000{,}000$ & $10$ & I \\
    \midrule
    \multirow{5}{*}{Final}
      & $100$           & $12$ & I & $1.0$ & $1.0$ & & $1.2,\ 1.0$ & $100$           & $10$ & I \\
      & $10{,}000$      & $\infty$ & I & $0.9$ & $0.8$ & & $1.0,\ 1.2$ & $10{,}000$      & $\infty$ & I \\
      & $10{,}000$      & $12$ & C & $0.8$ & $1.0$ & & $1.2,\ 1.5$ & $10{,}000$      & $1$  & C \\
      & $1{,}000{,}000$ & $12$ & I & $1.0$ & $0.8$ & & $0.8,\ 1.5$ & $1{,}000{,}000$ & $10$ & I \\
      & $1{,}000{,}000$ & $\infty$ & C & $0.9$ & $0.9$ & & $0.8,\ 1.5$ & $1{,}000{,}000$ & $\infty$ & C \\
    \bottomrule
  \end{tabular}
\end{table}

\section{Frontier-LLM tournament model list}
\label{app:glee_models}

The $13$ frontier LLMs in the round-robin tournament of~\citet{shapira2024glee}, grouped by provider.

\begin{table}[h]
  \centering
  \small
  \begin{tabular}{@{}l l@{}}
    \toprule
    Provider & Model \\
    \midrule
    Anthropic & Claude~3.5~Sonnet~(2024-10-22) \\
              & Claude~3.7~Sonnet~(2025-02-19) \\
    Google    & Gemini~1.5~Flash \\
              & Gemini~1.5~Pro \\
              & Gemini~2.0~Flash \\
              & Gemini~2.0~Flash-Lite \\
    Meta      & Llama-3.3-70B-Instruct \\
              & Llama-3.1-405B-Instruct-FP8 \\
    Mistral   & Mistral-Large-Latest \\
    OpenAI    & GPT-4o \\
              & GPT-4o-Mini \\
              & o3-Mini \\
    xAI       & Grok-2 (2024-12-12) \\
    \bottomrule
  \end{tabular}
\end{table}

All $13$ receive the same game-facing system prompt; the only variable across source agents is the underlying LLM.

\section{Hackathon competition details}
\label{app:hackathon}

This appendix expands on the held-out target population introduced in Section~\ref{sec:data}. The hackathon ran in December~$2025$ over four stages: three preliminary stages with progressively richer game configurations, and a final round in which the top six teams (selected by stage-$3$ payoff) competed for a \$$2{,}000$ prize pool. Game configurations released at each stage are listed in Appendix~\ref{app:configs}. To control API cost across $34$ teams playing thousands of games, we restricted participants to the \emph{Gemini~2.5~Flash} and \emph{Gemini~2.5~Flash-Lite} API surface; this choice is what makes the target population orthogonal to the frontier-LLM source on the variation axis (scaffolding instead of underlying LLM).

\paragraph{Released dataset vs.\ predictive cohort.}
We distinguish two cuts of the data. The publicly released dataset contains the $23$ final-round submissions for which complete code is available for audit. For the predictive evaluation reported in the main text, we additionally retain the decision logs of every $(\text{team}, \text{stage})$ agent that appeared at any stage of the competition: teams submitted revised versions between stages, so a single team contributes multiple distinct agents. This yields $91$ team-stage agents and $11{,}341$ accept/reject decisions over $4{,}921$ bargaining and negotiation games.

\section{Proposal prediction: model details}
\label{app:proposal}

This appendix documents the model choices specific to the regression task. Aggregate numbers appear in Table~\ref{tab:results_proposal}.

\paragraph{Task and inverse normalisation.} For each two-player game in which the target agent plays as proposer, and for each round $r{\geq}2$, we predict the offer the proposer will submit--equivalently, the offer its next opponent will face. Training is on a scale-normalised target so that configurations with different monetary scales live on the same axis; the dollar amount a stakeholder would see is recovered from the regressor output by a closed-form inverse. For bargaining, the normalised target is the proposer's own share of the divided sum ($\text{self\_gain}/(\text{self\_gain}+\text{other\_gain}) \in [0,1]$); inverting the normalisation returns the dollar offer to the opponent as $(1-\hat{y}) \cdot M$, where $M$ is the configuration's total to split. For negotiation, the normalised target is the proposer's price divided by a configuration-specific scale constant $S$ (the reference price defined per configuration); inverting returns $\hat{y} \cdot S$, the dollar price the opponent is being offered. Normalised values can exceed $1$ when the proposer's price sits above the nominal scale. Round $1$ is excluded because it has no prior-round state to condition on.

\paragraph{Task-oriented prompt suffix.} The reconstructed decision-time prompt used by the response-prediction model ends with \texttt{\{"decision":~"}, which primes the Observer for an \texttt{accept}/\texttt{reject} token. That suffix is uncorrelated with the proposal regression target. We therefore swap the suffix for a task-matched one that orients the Observer toward the proposer's own next offer:
\begin{itemize}
  \item Bargaining proposal: \texttt{Offer:~\{proposer\_name\}\_gain:~\$} — primes for the proposer's own dollar gain, which is our target variable for that family.
  \item Negotiation proposal: \texttt{Offer:~\$} — primes for the proposer's price amount.
\end{itemize}
This is the only change at Observer-extraction time. The response-prediction suffix is unchanged from the main model.

\paragraph{Feature stacks.} The proposal-prediction comparison in Table~\ref{tab:results_proposal} contrasts the same two stacks used for response prediction:
\begin{itemize}
  \item \emph{Game+text features baseline:} the proposer-side game-feature schema (round index, horizon, discount factors, per-side valuations, prior-round offers and decisions), the dialogue representation, and the agent-identity indicator.
  \item \emph{Observer-augmented model:} the same stack plus the Observer hidden-state representation. This is the row reported under each Observer model in Table~\ref{tab:results_proposal}.
\end{itemize}
The primary comparison is the Observer's marginal contribution on top of the game+text features baseline.

\paragraph{Cohort filters.} We include hackathon agents with at least $30$ round-${\geq}2$ proposer decisions and target standard deviation $\geq\!0.02$; bargaining passes $78$ agents, negotiation passes $20$.

\paragraph{Evaluation.} Same protocol as response prediction: cross-population transfer from the frontier-LLM tournament, training rows capped at $3{,}000$ (balanced across source agents), test rows capped at $500$ per cell, game-level splits, $K{\in}\{0,2,4,8,16\}$, $5$ seeds, TabPFN~v2.6 in regressor mode.

\paragraph{Why median.} Per-agent $R^{2}$ in this regression task is heavy-tailed: for some agents, predictions extrapolate to extreme values on a small subset of configurations, driving per-agent $R^{2}$ to large negative values. Means over agents are dominated by these tail cases, while medians are stable. We therefore report medians in Table~\ref{tab:results_proposal} and in the primary discussion of proposal-prediction results.

\section{Provider replication of the logits-vs-hidden-states gap}
\label{app:providers}

This appendix supports the robustness paragraph of Section~\ref{sec:results:robustness}. We test whether the gap between an Observer's direct logits and its hidden-state representation depends on the choice of provider. The comparison is run in the cross-population transfer protocol of Section~\ref{sec:results} (frontier-LLM source population, hackathon target population, $K{=}16$, TabPFN classifier, $5$ seeds), so the contrast is purely between two read-outs of the same Observer under the same evaluation setup used for the main results. The cohort matches the main response-prediction evaluation: $72$ bargaining and $39$ negotiation hackathon agents.

\paragraph{Logits vs.\ hidden states.}
Table~\ref{tab:signal_gap} isolates the effect for the Gemma-2-2B Observer. Logits alone, whether passed through a TabPFN classifier or read directly as $p(\text{accept})$, are far weaker than the Observer hidden-state representation. The game+text features baseline is already strong; adding the logit scalar yields only a marginal improvement, while adding hidden states yields a clearly larger one. Adding both logits and hidden states is a wash relative to hidden states alone. The pattern matches the Section~\ref{sec:results} takeaway: the predictive signal lives in the Observer's representation, not in its direct readout.

\begin{table}[h]
  \centering
  \caption{Matched comparison of logits vs.\ hidden states under cross-population transfer at $K{=}16$ with TabPFN, mean AUC over $5$ seeds (SE over $\text{agents}\times\text{seeds}$ in parentheses; $72$ bargaining, $39$ negotiation hackathon targets). Logits $= p(\text{accept})$ obtained by renormalising the Observer's next-token probabilities over the accept/reject verbalisers; hidden state $=$ Observer hidden-state representation averaged over Gemma-2-2B's upper-stack layer band (relative depth $0.6$--$0.9$).}
  \label{tab:signal_gap}
  \small
  \begin{tabular}{@{}l cc@{}}
    \toprule
    Input to the TabPFN classifier & Bargaining & Negotiation \\
    \midrule
    Logits alone                                     & .585\,\scriptsize{(.010)} & .464\,\scriptsize{(.015)} \\
    Observer hidden states alone                     & .701\,\scriptsize{(.008)} & .536\,\scriptsize{(.013)} \\
    Game-state features alone                        & .756\,\scriptsize{(.010)} & .827\,\scriptsize{(.014)} \\
    Game-state features + logits                     & .767\,\scriptsize{(.009)} & .831\,\scriptsize{(.014)} \\
    Game-state features + Observer hidden states     & \textbf{.793}\,\scriptsize{(.008)} & \textbf{.838}\,\scriptsize{(.013)} \\
    Game-state features + logits + Observer hidden states & .791\,\scriptsize{(.008)} & .835\,\scriptsize{(.013)} \\
    \bottomrule
  \end{tabular}
\end{table}

\paragraph{Replication across three providers.}
If the logit-vs-hidden-state gap were an artifact of the hackathon agents' underlying LLM matching the Observer's training pipeline, Observers from unrelated providers should fail to reproduce it. They do not. We replicate with Qwen3-1.7B~\citep{yang2025qwen3} (Alibaba) and Llama-3.2-1B~\citep{grattafiori2024llama} (Meta), neither of which shares a training pipeline or parent company with the hackathon agents' underlying LLM, and recover the same pattern (Table~\ref{tab:observer_robustness}). Game-state features plus Observer hidden states sit in a tight band across providers (bargaining $\in[0.76,0.79]$, negotiation $\in[0.84,0.86]$). Direct $p(\text{accept})$ AUC is more variable across providers, especially in negotiation: Gemma and Llama remain near chance ($\le 0.45$) while Qwen3 reaches $0.717$, an outlier we report transparently rather than aggregate over. The hidden-state-augmented predictions are consistent across providers; the direct readout is not.

\begin{table}[h]
  \centering
  \caption{Logits vs.\ hidden states across Observer providers, cross-population transfer at $K{=}16$. Logits columns report direct $p(\text{accept})$ AUC of the Observer (no classifier fit; values can fall below chance because the Observer's preferred direction is not enforced to align with acceptance). Hidden-state columns use game-state features plus the Observer hidden-state representation (averaged over the upper-stack layer band, relative depth $0.6$--$0.9$) under the same TabPFN classifier as Table~\ref{tab:signal_gap}.}
  \label{tab:observer_robustness}
  \small
  \begin{tabular}{@{}ll cc cc cc@{}}
    \toprule
    & & & \multicolumn{2}{c}{\emph{Logits alone}} & \multicolumn{2}{c}{\emph{Game + Hidden states}} \\
    \cmidrule(lr){4-5} \cmidrule(l){6-7}
    Observer & Provider & Params & Barg & Nego & Barg & Nego \\
    \midrule
    Gemma-2-2B   & Google  & 2.6B & .548 & .403 & .793 & .838 \\
    Qwen3-1.7B   & Alibaba & 1.7B & .607 & .717 & .785 & .863 \\
    Llama-3.2-1B & Meta    & 1.2B & .577 & .446 & .763 & .837 \\
    \bottomrule
  \end{tabular}
\end{table}

\section{Additional experimental details}
\label{app:details}

\paragraph{Hardware.}
All GPU experiments ran on an internal cluster with up to $2\times$ NVIDIA RTX A6000 ($48$\,GB each); individual jobs used a single GPU at a time. CPU-only steps (game-feature extraction and dialogue encoding with \texttt{all-MiniLM-L6-v2}) ran on the same hosts. The direct LLM-as-Predictor baseline (Section~\ref{sec:results} and Appendix~\ref{app:thinking_predictor}) is API-only and consumes no local GPU time.

\paragraph{Observer feature extraction.}
Decision-time hidden states for the three Observers (Gemma-2-2B, Qwen3-1.7B, Llama-3.2-1B) were extracted with TransformerLens \texttt{run\_with\_cache} over the reconstructed prompt for $\approx\!67$K games (bargaining and negotiation, both populations; $\approx\!200$K decisions). All upper-stack layers used in the layer sweep of Figure~\ref{fig:observer_layer} are cached in a single forward pass per game; the main Observer hidden-state representation is the average over each model's upper-stack layer band (relative depth $0.6$--$0.9$). Three suffix variants are extracted per Observer—response prediction (\texttt{\{"decision":\,"}), bargaining proposal, and negotiation proposal—because the suffix re-orients the Observer toward the task (Appendix~\ref{app:proposal}). The same forward pass also caches the per-decision next-token logits used in the logits-vs-hidden-states comparison of Appendix~\ref{app:providers}, so that comparison adds no extraction cost. Cumulative Observer extraction: $\approx\!80$\,A6000 GPU-hours.

\paragraph{Tabular evaluation.}
TabPFN~v2.6 was run with default settings; each evaluation cell uses up to $3{,}000$ source-balanced training rows together with the target's $K$-game decisions and at most $500$ test rows, taking $\approx\!10\text{--}30$\,s on a single A6000. The reported tabular budget covers all such cells across: the cross-population response and proposal tables (Tables~\ref{tab:results_response}--\ref{tab:results_proposal}); the layer sweep over each Observer's full stack (Figure~\ref{fig:observer_layer}), which dominates this budget because every layer is evaluated separately rather than averaged over the upper-stack band; and the logits-vs-hidden-states matched comparison and its three-provider replication (Appendix~\ref{app:providers}). Cumulative tabular evaluation: $\approx\!60$\,A6000 GPU-hours.

\paragraph{LLM-as-Predictor (API).}
The LLM-as-Predictor numbers in Tables~\ref{tab:results_response}--\ref{tab:results_proposal} use Gemini~2.5~Flash with \texttt{thinking\_budget=0}: $65$ bargaining and $33$ negotiation hackathon targets $\times$ $K \in \{0,2,4,8,16\}$ $\times$ $5$ seeds $\times$ up to $30$ test decisions per cell. Because no thinking tokens are generated, the per-call cost is much lower than the pilot. The thinking-on pilot in Appendix~\ref{app:thinking_predictor} used $4{,}010$ calls with \texttt{thinking\_budget=2000} at a cost of $\approx\!\$31$; we did not scale this configuration to the full table, which would have cost an estimated $\approx\!\$235$.

\paragraph{Total.}
The experiments reported in the paper consumed $\approx\!140$\,A6000 GPU-hours cumulatively (extraction $+$ tabular evaluation), plus the Gemini API budget above. Including preliminary and discarded runs—alternative Observer layers and pooling strategies, predictor sweeps that did not converge under the cross-population protocol, abandoned feature stacks, and exploratory hackathon-only protocols not reported in the final paper—the broader project consumed roughly $\approx\!400$\,GPU-hours.

\paragraph{Model hyperparameters.} TabPFN~v2.6 with default settings (no tuning). Observer hidden-state representations are averaged over each model's upper-stack layer band (relative depth $0.6$--$0.9$); layer counts per Observer are Gemma-2-2B ($26$), Qwen3-1.7B ($28$), and Llama-3.2-1B ($16$). Seeds $\{0,1,2,3,4\}$. Full feature-block specifications (game-feature schemas per family, dialogue construction, PCA dimensions, identity-column construction) are in Appendix~\ref{app:baseline_specs}.

\paragraph{Observer prompt suffixes.} The suffix appended after the dialogue history orients the Observer toward the decision to predict: \texttt{\{"decision":~"} for response prediction (the original GLEE accept/reject format), \texttt{Offer:~\$} for proposal prediction in negotiation, and \texttt{Offer:~\{proposer\_name\}\_gain:~\$} for proposal prediction in bargaining (proposer's own dollar gain).

\paragraph{Game-level split.} For each held-out target agent, at each $K$, we randomly select $K$ of the agent's games as the $K$-shot pool (all round-by-round decisions within those games enter the adaptation rows); the remaining games form the test set. Splitting at the game level ensures that rounds from the same game never appear on both sides of the split, and makes explicit that $K$ counts whole observed games--each with multiple decisions--rather than individual decisions.

\paragraph{Error bars.} Throughout the paper, $\pm$ values report standard errors of the per-$(\text{agent},\text{seed})$ AUC or $R^{2}$ values for each cell ($\sigma/\sqrt{N}$ with $N = N_\text{targets} \times 5$). The Observer-by-layer figure (Figure~\ref{fig:observer_layer}) reports paired $95\%$ confidence intervals as $1.96 \cdot \mathrm{SEM}$ on the per-$(\text{agent},\text{seed})$ deltas of full $-$ baseline, anchored at the baseline mean.

\section{Game+text features baseline: feature specifications}
\label{app:baseline_specs}

This appendix specifies the three feature blocks of the game+text features baseline (Section~\ref{sec:method}): game-state features, the dialogue representation, and the agent-identity indicator. The Observer hidden-state representation, appended in the Observer-augmented model, is documented in Appendix~\ref{app:details}.

\begin{figure}[h]
  \centering
  \includegraphics[width=\linewidth]{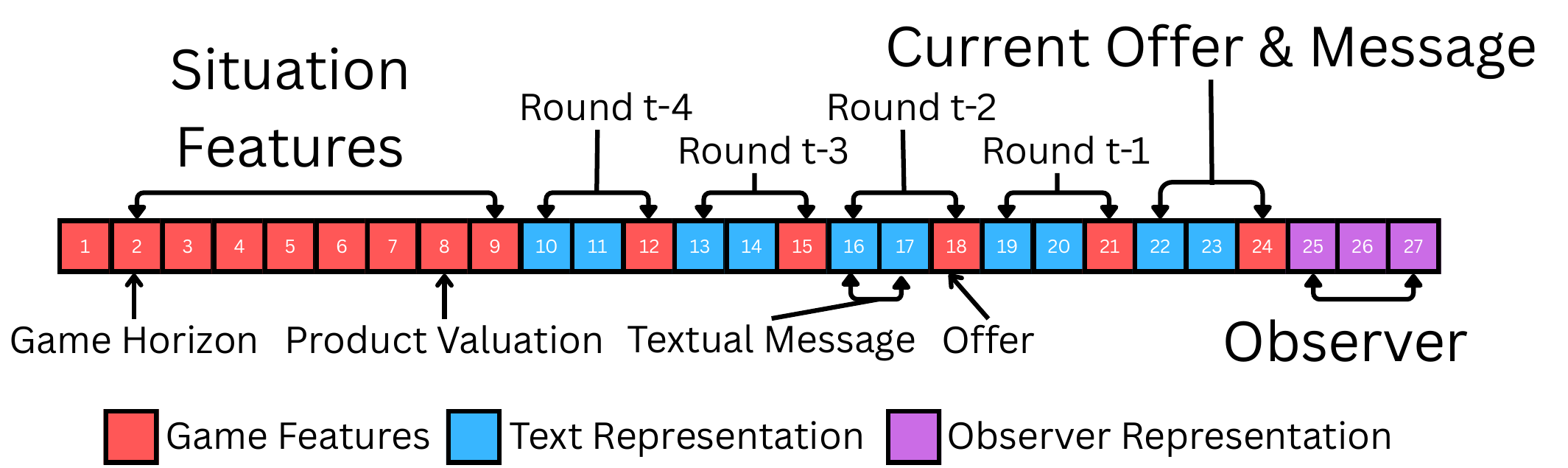}
  \caption{Schematic of the multimodal tabular row at a single decision point. The row concatenates the three feature modalities of Section~\ref{sec:method}: game-state features (red), the dialogue representation produced by the sentence encoder (blue), and the Observer hidden-state representation of the current decision-time state (purple). Game-state features are divided into configuration-level situation features (e.g., game horizon, product valuation) and per-round entries summarizing the last few rounds and the current offer; the dialogue representation contributes per-round textual entries. Cell counts are illustrative; actual modality dimensions and game-feature columns differ by game family (bargaining vs.\ negotiation).}
  \label{fig:feature_modalities}
\end{figure}

\paragraph{Game-state features -- bargaining ($24$ columns).}
At each responder decision at round $r$ we extract a fixed-schema vector from the publicly observable game state, with $N{=}5$ prior rounds of history. Missing entries (history slots before round $1$, or fields that do not apply in a particular configuration) are encoded as \texttt{NaN}; TabPFN handles \texttt{NaN} natively. The columns are:
\begin{itemize}
\item \emph{Configuration ($8$):} \texttt{round}, \texttt{max\_rounds}, \texttt{round\_frac} ($=\!r/\texttt{max\_rounds}$), \texttt{money} (the amount to divide $M$), \texttt{delta\_1} and \texttt{delta\_2} (the two players' per-round discount factors), \texttt{messages} (binary: free-text exchange allowed in this game), \texttt{complete\_info} (binary: parameters shared with both players or private).
\item \emph{Current offer ($5$):} \texttt{offer\_frac} ($=\!\texttt{responder\_gain}/(\texttt{proposer\_gain}{+}\texttt{responder\_gain})$), \texttt{responder\_gain} and \texttt{proposer\_gain} (the split in absolute units), \texttt{inflation\_loss\_1} ($=\!1{-}\texttt{delta\_1}^{r-1}$) and \texttt{inflation\_loss\_2} (cumulative discounting incurred up to round $r$).
\item \emph{History ($10$):} for $h{=}1,\dots,5$, \texttt{prev}$h$\texttt{\_offer\_frac} and \texttt{prev}$h$\texttt{\_decision}: the proposer's split and responder's accept/reject at round $r{-}h$.
\item \emph{Family indicator ($1$):} \texttt{family\_idx} (constant within bargaining; included for cross-family compatibility of the schema).
\end{itemize}

\paragraph{Game-state features -- negotiation ($25$ columns).}
At each responder decision we extract:
\begin{itemize}
\item \emph{Configuration ($8$):} \texttt{round}, \texttt{max\_rounds}, \texttt{round\_frac}, \texttt{sv} (seller's reservation valuation), \texttt{bv} (buyer's reservation valuation), \texttt{product\_price\_order} (typical-price scale used to order configurations), \texttt{messages}, \texttt{complete\_info}.
\item \emph{Outside-option references ($2$):} \texttt{seller\_outside} ($=\!\texttt{sv}\cdot\texttt{product\_price\_order}$) and \texttt{buyer\_outside} ($=\!\texttt{bv}\cdot\texttt{product\_price\_order}$): the absolute payoffs of the outside option for each player.
\item \emph{Current offer ($4$):} \texttt{price} (the proposed sale price), \texttt{offer\_frac} ($=\!\texttt{price}/\texttt{product\_price\_order}$), \texttt{offer\_vs\_buyer\_outside} ($=\!\texttt{price}/\texttt{buyer\_outside}$, the relative cost compared to the buyer's outside option), \texttt{rounds\_remaining}.
\item \emph{History ($10$):} \texttt{prev}$h$\texttt{\_offer\_frac} and \texttt{prev}$h$\texttt{\_decision} for $h{=}1,\dots,5$.
\item \emph{Family indicator ($1$):} \texttt{family\_idx}.
\end{itemize}

\paragraph{Dialogue representation.}
For each responder decision at round $r$, we collect every \texttt{message} field exchanged within round $r$ in the game log (the proposer's message accompanying the offer, plus any responder-side message in the same round), concatenate them into a single string with single-space separators, and encode the string with the \texttt{sentence-transformers/all-MiniLM-L6-v2} sentence encoder, yielding a $384$-dimensional vector. When the round contains no messages we use the placeholder \texttt{"Round~}$r$\texttt{"} so the representation is always defined. The $384$-dimensional vectors are PCA-reduced to $5$ dimensions; the projection is fit on the training pool of each evaluation cell and applied unchanged to the test pool. The choice of $5$ is intentionally low: a value such as $30$ would be a more natural default for MiniLM, but $5$ performed slightly better for the game+text features baseline at $K{=}0$ in pilot runs, and we keep the same value for the Observer-augmented model so that any difference between the two is not attributable to dialogue-representation capacity.

\paragraph{Agent-identity indicator.}
Let $\mathcal{S}$ denote the source agents in the training pool and $t$ the held-out target agent. The agent-identity indicator is a one-hot vector of dimension $|\mathcal{S}|+1$ over $\mathcal{S}\cup\{t\}$, with a single $1$ marking the agent whose decision is being predicted. At $K{=}0$, every training row lies in $\mathcal{S}$ (so the $t$-column is always zero in training) and every test row activates the $t$-column; the column therefore separates train from test deterministically and adds no within-test signal. At $K{>}0$, the $K$ adaptation games of $t$ enter the training pool and activate the $t$-column there, which gives the tabular predictor a within-target anchor when it predicts on the held-out test games of $t$.

\paragraph{PCA fitting.}
All PCA projections in the model (dialogue representation, Observer hidden states) are fit on the training pool of each evaluation cell and then applied unchanged to the test pool. No test-row information enters PCA fitting at any $K$.

\section{Thinking-budget pilot for the LLM-as-Predictor baseline}
\label{app:thinking_predictor}

The main LLM-as-Predictor table uses a full run with \texttt{thinking\_budget=0}: $65$ bargaining and $33$ negotiation hackathon target agents, $K\in\{0,2,4,8,16\}$, and up to $30$ test decisions per cell, matching the seed protocol of Section~\ref{sec:setup}. To check whether this underpowers the direct LLM-as-Predictor baseline, we ran a stratified pilot on $5$ target agents per family with \texttt{thinking\_budget=2000}. The pilot cost was approximately \$31 for $4{,}010$ API calls, with thinking tokens accounting for most of the cost; scaling the same configuration to the full table was estimated at roughly \$235, so we did not run it as a main experiment.

\begin{table}[h]
  \centering
  \caption{Thinking-budget pilot on the same $5$ target agents per family and matched cells. The rightmost column reports our cross-population response-prediction model (game+text features + Observer hidden-state representation) restricted to the pilot cells.}
  \label{tab:thinking_pilot}
  \small
  \begin{tabular}{@{}l l ccc@{}}
    \toprule
    Family & $K$ & LLM-as-Predictor (budget 0) & LLM-as-Predictor (budget 2000) & Observer-augmented model \\
    \midrule
    Barg & 0 & .628 & .656 & .672 \\
    Barg & 8 & .792 & .860 & .857 \\
    Barg & 16 & .733 & .806 & \textbf{.881} \\
    \midrule
    Nego & 0 & .492 & \textbf{.700} & .623 \\
    Nego & 8 & .645 & .636 & \textbf{.771} \\
    Nego & 16 & .720 & .642 & \textbf{.750} \\
    \bottomrule
  \end{tabular}
\end{table}

The result is mixed rather than a monotone strengthening of the LLM-as-Predictor. Thinking improves some low-$K$ cells, including negotiation at $K{=}0$, but hurts negotiation at high $K$ on this subset. The pilot does not overturn the main comparison: a stronger direct LLM-as-Predictor can improve isolated cells, especially at low $K$, but the Observer-augmented model remains stronger at $K{=}16$ in both families.

\section{Broader impacts}
\label{app:impact}

Predicting unfamiliar agents' decisions has dual-use implications. Constructive uses include mechanism design, planning under counterpart uncertainty, and prediction-aware agents in language-mediated commerce that adapt to a partner's behaviour without sharing private state. The same capability could be misused for adversarial counterpart modelling--e.g., extracting concession patterns to exploit them--deployed without disclosure to the counterparty. As predictive models of language-based agents become more accurate, transparency norms about whether counterparts are being modelled, analogous to consumer-facing disclosures, become an important complement to the technical work.

\end{document}